\documentclass{bioinfo}

\usepackage{graphicx}
\usepackage{tabularx,ragged2e}
\usepackage{multirow}
\usepackage{amsmath,amssymb,amsthm}
\usepackage{txfonts}
\usepackage{algorithm}
\usepackage{algorithmic}
\usepackage{subfig}
\usepackage{caption}
\usepackage{url}
\usepackage{color}
\usepackage{natbib}

\copyrightyear{2017}
\pubyear{2017}

\begin{document}
\firstpage{1}

\newcommand{\red}[1]{\textcolor{black}{#1}}

\renewcommand\footnotemark{}

\title[Deep learning guided Bayesian inference for super-resolution fluorescence microscopy]{DLBI: Deep learning guided Bayesian inference for structure reconstruction of super-resolution fluorescence microscopy}
\author[Li et al.]{Yu Li$^{1,\dagger}$, Fan Xu$^{2,\dagger}$\thanks{$^{\dagger}$These authors contributed equally to this work.}, Fa Zhang$^{2}$, Pingyong Xu$^{3}$, Mingshu Zhang$^3$, Ming Fan$^4$, Lihua Li$^4$, Xin Gao$^{1,*}$, Renmin Han$^{1,*}$\thanks{$^{*}$All correspondence should be addressed to Xin Gao (xin.gao@kaust.edu.sa) and Renmin Han (renmin.han@kaust.edu.sa).} }
\address{$^1$King Abdullah University of Science and Technology (KAUST), Computational Bioscience Research Center (CBRC), Computer, Electrical and Mathematical Sciences and Engineering (CEMSE) Division, Thuwal, 23955-6900, Saudi Arabia. $^2$High Performance Computer Research Center, Institute of Computing Technology, Chinese Academy of Sciences, Beijing 100190, China. $^3$Key Laboratory of RNA Biology, Institute of Biophysics, Chinese Academy of Sciences, Beijing 100101, China. $^4$Institute of Biomedical Engineering and Instrumentation, Hangzhou Dianzi University, Hangzhou, 310018, China.
}


\maketitle

\begin{abstract}

\section{Motivation:}
Super-resolution fluorescence microscopy, with a resolution beyond the diffraction limit of light, has become an indispensable tool to directly visualize biological structures in living cells at a nanometer-scale resolution. Despite advances in high-density super-resolution fluorescent techniques, existing methods still have bottlenecks, including extremely long execution time, artificial thinning and thickening of structures, and lack of ability to capture latent structures.

\section{Results:}
Here we propose a novel deep learning guided Bayesian inference approach, DLBI, for the time-series analysis of high-density fluorescent images. Our method combines the strength of deep learning and statistical inference, where deep learning captures the underlying distribution of the fluorophores that are consistent with the observed time-series fluorescent images by exploring local features and correlation along time-axis, and statistical inference further refines the ultrastructure extracted by deep learning and endues physical meaning to the final image. In particular, our method contains three main components. The first one is a simulator that takes a high-resolution image as the input, and simulates time-series low-resolution fluorescent images based on experimentally calibrated parameters, which provides supervised training data to the deep learning model. The second one is a multi-scale deep learning module to capture both spatial information in each input low-resolution image as well as temporal information among the time-series images. And the third one is a Bayesian inference module that takes the image from the deep learning module as the initial localization of fluorophores and removes artifacts by statistical inference. Comprehensive experimental results on both real and simulated datasets demonstrate that our method provides more accurate and realistic local patch and large-field reconstruction than the state-of-the-art method, the 3B analysis, while our method is more than two orders of magnitude faster.

\section{Availability:}
The main program is available at \url{https://github.com/lykaust15/DLBI}.

\end{abstract}

\vspace{-0.7cm}
\section{Introduction}
Fluorescence microscopy with a resolution beyond the diffraction limit of light (i.e., super-resolution) has played an important role in biological sciences. The application of super-resolution fluorescence microscope techniques to living-cell imaging promises dynamic information on complex biological structures with nanometer-scale resolution.

Recent development of fluorescence microscopy takes advantages of both the development of optical theories and computational methods. Living cell stimulated emission depletion (STED) \citep{hein2008stimu}, reversible saturable optical linear fluorescence transitions (RESOLFT) \citep{a2007wide}, and structured illumination microscopy (SIM) \citep{gustafsson2005nonlinear} mainly focus on the innovation of instruments, which requires sophisticated, expensive optical setups and specialized expertise for accurate optical alignment. The time-series analysis based on localization microscopy techniques, such as photoactivatable localization microscopy (PALM) \citep{hess2006ultra} and stochastic optical reconstruction microscopy (STORM) \citep{rust2006sub}, is mainly based on the computational methods, which build a super-resolution image from the localized positions of single molecules in a large number of images. Though compared with STED, RESOLFT and SIM, PALM and STORM do not need specialized microscopes, the localization techniques of PALM and STORM require the fluorescence emission from individual fluorophores to not overlap with each other, leading to long imaging time and increased damage to live samples \citep{lippincott2009putting}. More recent methods \citep{holden2011daostorm,huang2011simultaneous,quan2011high,zhu2012faster} alleviate the long exposure problem by developing multiple-fluorophore fitting techniques to allow relatively dense fluorescent data, but still do not solve the problem completely.

Bayesian-based time-series analysis of high-density fluorescent images \citep{cox2012bayesian,xu2015bayesian,xu2016live} further pushes the limit. By using data from overlapping fluorophores as well as information from blinking and bleaching events, it extends the super-resolution imaging to the large-field imaging of living cells. Despite its potential to resolve ultrastructures and fast cellular dynamics in living cells, several bottlenecks still remain. The state-of-the-art methods, such as Bayesian analysis of the blinking and bleaching (i.e., the 3B analysis) \citep{cox2012bayesian}, are computationally expensive, and may cause artificial thinning and thickening of structures due to local sampling. Significant improvements on runtime and accuracy have been achieved by single molecule-guided Bayesian localization microscopy (SIMBA) \citep{xu2016live} with the introduction of dual-channel fluorescent imaging and single molecule-guided Bayesian inference. However, the enhanced process is severely limited by the specialized class of proteins.

Deep learning has accomplished great success in various fields, including super-resolution imaging \citep{ledig2016photo,kim2016accurate,lim2017enhanced}. Among different deep learning architectures, the generative adversarial network (GAN) \citep{goodfellow2014} achieved the state-of-the-art performance on single image super-resolution (SISR) \citep{ledig2016photo}. However, there are two fundamental differences between the SISR and super-resolution fluorescence microscopy. First, the input of SISR is a downsampled (i.e., low-resolution) image of a static high-resolution image and the expected output is the original image, whereas the input of super-resolution fluorescence microscopy is a time-series of low-resolution fluorescent images and the output is the high-resolution image containing estimated locations of the fluorophores (i.e., the reconstructed structure). Second, the nature of SISR ensures that there are readily a huge amount of existing data to train deep learning models, whereas for fluorescence microscopy, there are only limited time-series datasets. Furthermore, most of these datasets do not have the ground-truth high-resolution images, which make supervised deep learning infeasible.

In this paper, we propose a novel deep learning guided Bayesian inference framework, DLBI, for structure reconstruction of high-resolution fluorescent microcopy. Our framework combines the strength of stochastic simulation, deep learning and statistical inference. To our knowledge, this is the first deep learning-based super-resolution fluorescent microscopy method. In particular, the stochastic simulation module simulates time-series low-resolution images from high-resolution images based on experimentally calibrated parameters of fluorophores and stochastic modeling, which provides supervised training data for deep learning models. The deep learning module takes the simulated time-series low-resolution images as inputs, captures the underlying distribution that generates the ground-truth super-resolutions images by exploring local features and correlation along time-axis of the low-resolution images, and outputs a predicted high-resolution image. To achieve this goal, we develop a generative adversarial network (GAN) in which a generator network and a discriminator network contest with each other. The generator network tries to learn the distribution of the high-resolution images in a multi-scale manner, whereas the discriminator network tries to discriminate the ground-truth images and the images produced by the generator network. In order to capture the deep features in the images, we further ease the degradation issue by integrating residual networks \citep{RN7} into our GAN model, where degradation means that stacking more network layers does not lead to better accuracy. The high-resolution image produced by the deep learning module is often very close to the ground-truth image. However, it can still contain some artifacts, and more importantly, lacks the physical meaning. Thus, we develop the Bayesian inference module to take the predicted high-resolution image from deep learning, run Bayesian inference from the initial locations of fluorophores in the predicted image, and predict a more accurate high-resolution image.

Comprehensive experimental results on two simulated and three real-world datasets demonstrate that DLBI provides more accurate and realistic local-patch as well as large-field reconstruction than the state-of-the-art method, the 3B analysis \citep{cox2012bayesian}. Meanwhile, our method is more than 100 times faster than the 3B analysis.

\vspace{-0.5cm}
\section{Materials and Methods}
\begin{figure}[!hpbt]
\centering
\vspace{-1.8em}
\includegraphics[width=0.40 \textwidth,type=png,ext=.png,read=.png]{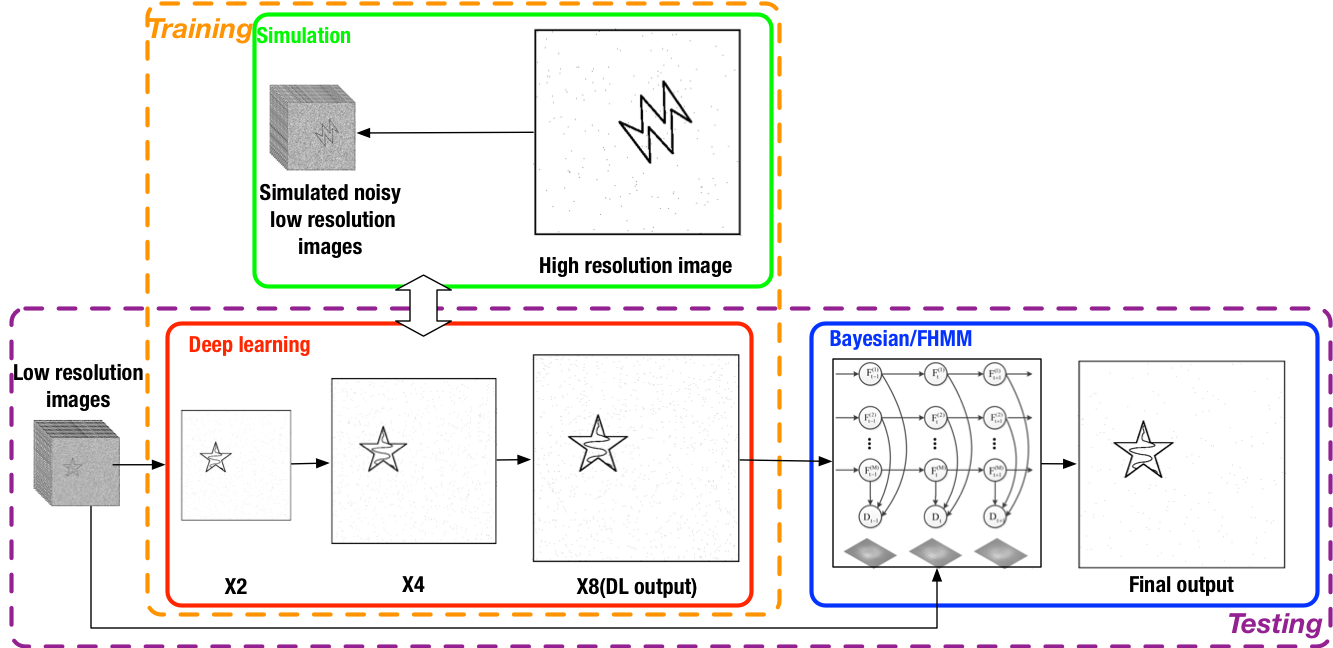}
\vspace{-1em}
\caption{The overall workflow of DLBI. The three modules are shown in solid boxes: the simulation module (green), the deep learning module (red), and the Bayesian inference module (blue). The training (orange) and testing (purple) procedures are shown in dashed boxes. For training, time-series low-resolution images simulated from the simulation module are used to train the deep learning module in a multi-scale manner, from 2X, 4X, to 8X resolution. For testing, given certain time-series low-resolution images, the deep learning module predicts the 2X to 8X super-resolution images, and the Bayesian inference module takes the predicted 8X image (which contains artifacts) and produces the final high-resolution image.}
\label{fig:workflow}
\vspace{-3.5em}
\end{figure}

As shown in Fig. \ref{fig:workflow}, DLBI contains three modules: (i) stochastic simulation (Section 2.1), (ii) deep neural networks (Section 2.2), and (iii) Bayesian inference (Section 2.3).

Although deep learning has proved its great superiority in various fields, it has not been used for fluorescent microscopy image analysis. One of the possible reasons is the lack of supervised training data, which means the number of time-series low-resolution image datasets is limited and even for the existing datasets, the ground-truth high-resolution images are often unknown. Here, a stochastic simulation based on the experimentally calibrated parameters is designed to solve this issue, without the need of collecting a massive amount of real fluorescent images. This empowers our deep neural networks to effectively learn the latent structures under the low-resolution, high-noise and stochastic fluorescing conditions. The primitive super-resolution images produced by deep neural networks still contain artifacts and lack physical meaning, we finally develop a Bayesian inference module based on the mechanism of fluorophore switching to produce high-confident images.

Our method combines the strength of deep learning and statistical inference, where deep learning captures the underlying distribution that generates the training super-resolution images by exploring local features and correlation along time-axis, and statistical inference removes artifacts and refines the ultrastructure extracted by deep learning, and further endues physical meaning to the final image.

\vspace{-0.3cm}
\subsection{The stochastic simulation module}
The input of our simulation module is a high-resolution image that depicts the distribution of the fluorophores and the output is a time-series of low-resolution fluorescent images with different fluorescing states. We refer the readers to Section S1 for terminologies in fluorescence microscopy.

\begin{figure}[!b]
\centering
\vspace{-2em}
\includegraphics[width=0.36 \textwidth,type=png,ext=.png,read=.png]{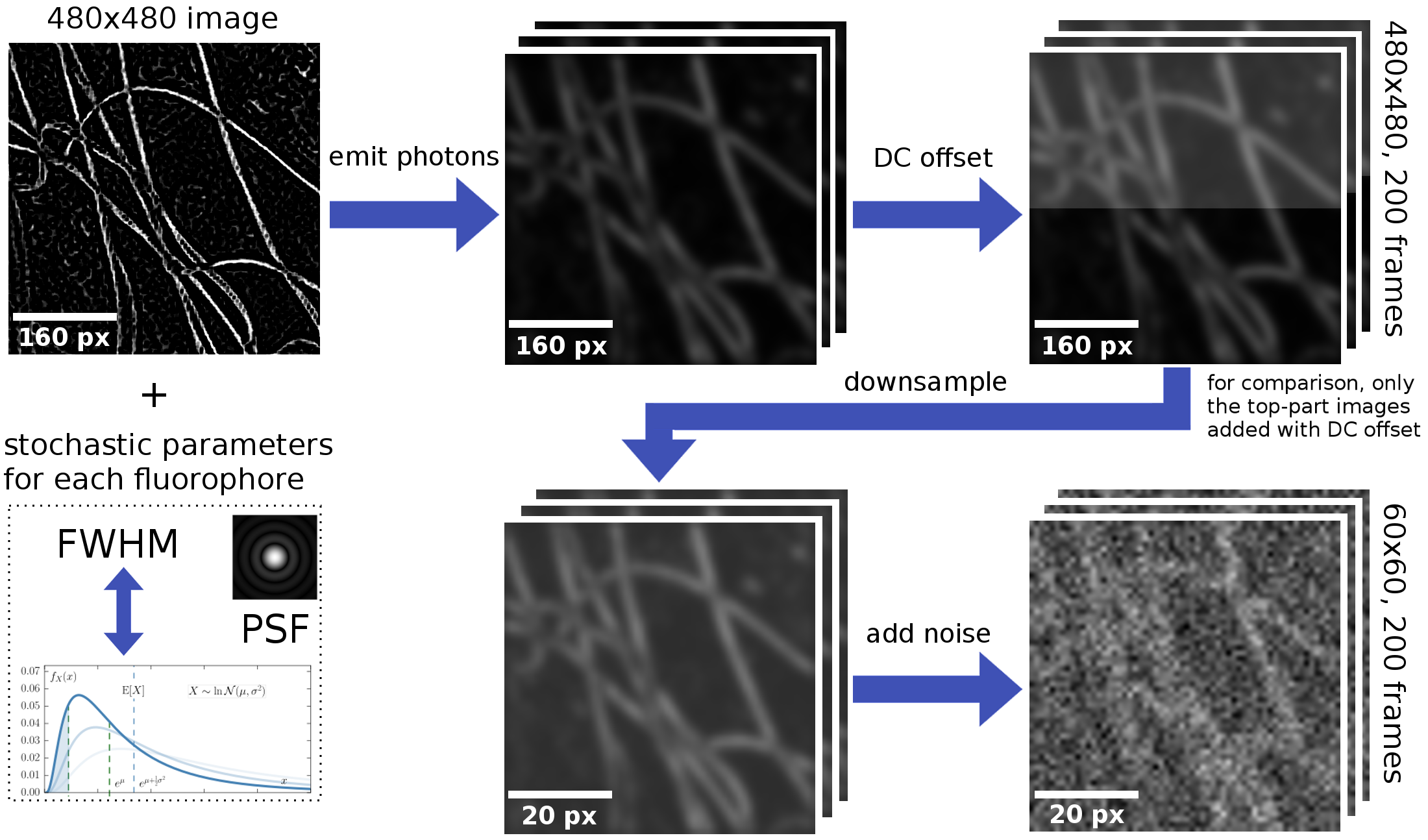}
\vspace{-1em}
\caption{The workflow of stochastic simulation. Firstly, a high-resolution image is inputted as the distribution and density of fluorophores. Then, the emitting of photons is simulated based on the stochastic parameters for each time frame. A random background (DC offset) is added to each image. The images are then downsampled to low-resolution and noise is added, which results in a time-series of low-resolution images. }
\label{fig:simu}
\vspace{-3.3em}
\end{figure}

In our simulation, Laplace-filtered natural images and sketches are used as the ground-truth high-resolution images that contain the fluorophore distribution. If a gray-scale image is given, the depicted shapes are considered as the distribution of fluorophores and each pixel value on the image is considered as the density of fluorophores at the location. We then create a number of simulated fluorophores that are distributed according to the distribution and the densities. For each fluorophore, it switches according to a Markov model, i.e., among states of emitting (activated), not emitting (inactivated), and bleached. The emitting state means that the fluorophore emits photons and a spot according to the point spread function (PSF) is depicted on the canvas. All the spots of the emitting fluorophores thus result in a high-resolution fluorescent image. Applying the Markov model on the initial high-resolution image generates a time-series of high-resolution images. After adding the background to the high-resolution images, they are downsampled to low-resolution images and noise is finally added. Fig. \ref{fig:simu} summarizes the stochastic simulation procedure.

Here, the success of simulation relies on three factors: (i) the principal of the linear optical system, (ii) experimentally calibrated parameters of fluorophores, and (iii) stochastic modeling.

\vspace{-0.3cm}
\subsubsection{Linear optics}
A fluorescence microscope is considered as a linear optical system, in which the superposition principle is valid, i.e., Image(Obj1 + Obj2) = Image(Obj1) + Image(Obj2). The behavior of fluorophores is considered invariant to mutual interaction. Therefore, for high-density fluorescent images, the pixel density can be directly calculated from the light emitted from its surrounding fluorophores. 

When a fluorophore is activated, an observable spot can be recorded by the sensor, the shape of which is called the point spread function (PSF). Considering the limitation of sensor capability, the PSF of an isotropic point source is often approximated as a Gaussian function:
\begin{footnotesize}
\begin{eqnarray}
   I(x, y) = I_0\exp(-\frac{1}{2\sigma^2}((x-x_0)^2 + (y-y_0)^2)),\label{eq:psf}
\end{eqnarray}
\end{footnotesize}
where $\sigma$ is calculated from the fluorophore in the specimen that specifies the width of the PSF, $I_0$ is the peak intensity and is proportional to the photon emission rate and the single-frame acquisition time, $(x_0, y_0)$ is the location of the fluorophore.

While PSF describes the shape, the full width at half maximum (FWHM) describes the distinguishability. It is defined to be the half width of the maximum amplitude of PSF. If PSF is modeled as a Gaussian function, the relationship between FWHM and $\sigma$ is given by
\begin{footnotesize}
\begin{eqnarray}
   \mathrm{FWHM} = 2\sqrt{2\ln2}\;\sigma \approx 2.355\;\sigma. \label{eq:fwhm}
\end{eqnarray}
\vspace{-0.2cm}
\end{footnotesize}
\vspace{-0.2cm}

Considering the probability of linear optics, a high-density fluorescent image is composed by PSFs of the fluorophores.

\vspace{-0.3cm}
\subsubsection{Calibrated parameters of fluorophores}
In most imaging systems, the characteristics of a fluorescent protein can be calibrated by experimental techniques. With all the calibrated parameters, it is not difficult to describe and simulate the fluorescent switching of a specialized protein.

\begin{figure}[!hpbt]
\centering
\vspace{-1.5em}
\includegraphics[width=0.31 \textwidth,type=png,ext=.png,read=.png]{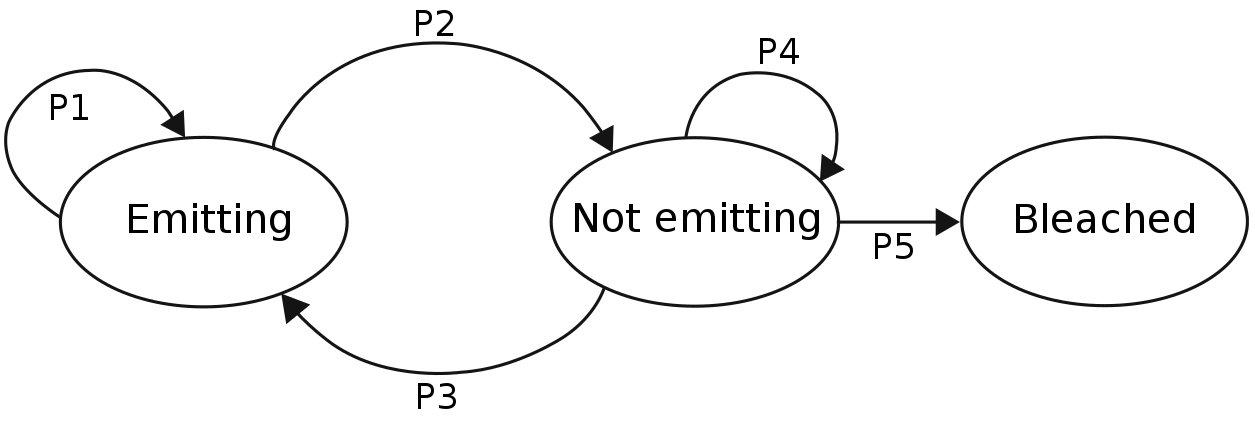}
\vspace{-1.5em}
\caption{The Markov model describing state transition of a fluorophore.}
\label{fig:trans}
\vspace{-4em}
\end{figure}

The first characteristic of a fluorophore is its switching probability. A fluorophore always transfers among three states, emitting, not emitting and bleached, which can be specified by a Markov model (Fig. \ref{fig:trans}). If the fluorophore transfers from not emitting to bleached, it will not emit any photon anymore. As linear optics, each fluorophore's transitions are assumed to be independent.

The second characteristic of a fluorophore is its PSF. When a real-world fluorophore is activated, the emitted photons and its corresponding PSF will not stay unchanged over time. The stochasticity of the PSF and photon strength describes the characteristics of a fluorescent protein. To simulate the fluorescence, we should not ignore these properties. Fortunately, the related parameters can be well-calibrated. The PSF and FWHM of a fluorescent protein can be measured in low molecule density. In an instrument for PALM or STORM, the PSF of the microscope can be measured by acquiring image frames, fitting the fluorescent spots parameter, normalizing and then averaging the aligned single-molecule images. The distribution of FWHM can be obtained from statistical analysis. The principle of linear optics ensures that the parameters measured in single-molecule conditions is also applicable to high-density conditions.

In our simulation, a log-normal distribution \citep{cox2012bayesian,zhu2012faster} is used to approximate the experimentally measured single fluorophore photon number distribution. Firstly, a table of fluorophore's experimentally calibrated FWHM parameters is used to initialize the PSF table in our simulation, according to Eq.\ref{eq:psf} and Eq.\ref{eq:fwhm}. Then for each fluorophore recorded in the high-resolution image, the state of the current image frame is calculated according to the transfer table [P1, P2, P3, P4, P5] (Fig. \ref{fig:trans}) and a random PSF shape is produced if the corresponding fluorophore is at the ``emitting'' state. This procedure is repeated for each fluorophore, which results in the final fluorescent image.

\vspace{-0.3cm}
\subsubsection{Stochastic modeling}
The illumination of real-world objects is different at different time. In general, the illumination change of real-world objects can be suppressed by high-pass filtering with a large Gaussian kernel. However, this operation will sharpen the random noise and cannot remove the background (or DC offset\footnote{DC offset, DC bias or DC component denotes the mean value of a signal. If the mean amplitude is zero, there is no DC offset. For most microscopy, the DC offset can be calibrated but cannot be completely removed.}). To make our simulation more realistic, several stochastic factors are introduced. First, for a series of simulated fluorescent images, a background value calculated from the multiplication between a random strength factor and the average image intensity is added to the fluorescent images to simulate the DC offset. For the same time-series, the strength factor remains unchanged but the background strength changes with the image intensity. Second, the high-resolution fluorescent image is downsampled and random Gaussian noise is added to the low-resolution image. Here, the noise is also stochastic for different time-series and close to the noise strength that is measured from the real-world microscopy.

The default setting of our simulation takes a $480\times480$ pixel high-resolution image as the input and simulates 200 frames of $60\times60$ pixel (i.e., $8\times$ binned) low-resolution images.

\vspace{-0.3cm}
\subsection{The deep learning module}
\begin{figure}[!hpbt]
\centering
\vspace{-2em}
\includegraphics[width=0.41 \textwidth,type=png,ext=.png,read=.png]{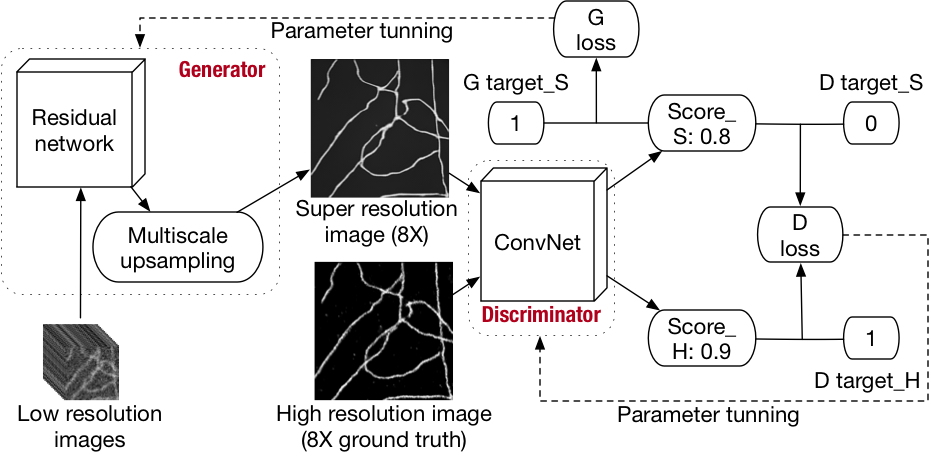}
\vspace{-1.em}
\caption{Overview of our deep learning framework. There are two main components of the model, the generator network ($\mathbf{G}$) and the discriminator network ($\mathbf{D}$). $\mathbf{G}$ is used to convert the time-series noisy, low-resolution images into a noise-free, super-resolution image while $\mathbf{D}$ is used to distinguish the ground-truth high-resolution image from the one produced by $\mathbf{G}$. $\mathbf{G}$ and $\mathbf{D}$ are designed to contest each other, which are trained simultaneously. As the training goes on, both $\mathbf{G}$'s ability to generate better super-resolution images and $\mathbf{D}$'s ability to distinguish the generated images are improved, which results in more and more similar images to the ground-truth from $\mathbf{G}$. During testing, only $\mathbf{G}$ is used.} 
\label{fig:gan}
\vspace{-3.5em}
\end{figure}

We build a deep residual network under the generative adversarial network (GAN) framework \citep{goodfellow2014,ledig2016photo} to estimate the primitive super-resolution image $I^{SR}$ (the latent structure features) from time-series of low-resolution fluorescent images $ \mathcal{T}=\{I^{FL}_{k}\}_{k=1,...,K}$. Instead of building just one generative model, our approach builds a pair of models, a generator model, $\mathbf{G}$, which produces the estimation of the underling structure of the training images, and a discriminator model, $\mathbf{D}$, which is trained to distinguish the reconstructed super-resolution image from the ground-truth one. Fig. \ref{fig:gan} demonstrates the overview of our deep learning framework.

\vspace{-0.3cm}
\subsubsection{Basic concepts}
The goal of training a generator neural network is to obtain the optimized parameters, $\theta_G$, for the generating function, $G$, with the minimum difference between the output super-resolution image, $I^{SR}$, and ground-truth, $I^{HR}$:
\begin{footnotesize}
\begin{eqnarray}
  \hat{\theta}_G=\mathop{\arg\min}\limits_{\theta_G}\frac{1}{N}\sum_{n=1}^{N}{l^{SR}(G(\mathcal{T}_n, \theta_G), I^{HR}_n)},
\end{eqnarray}
\end{footnotesize}
where $G(\mathcal{T}_n, \theta_G)$ is the generated super-resolution image by $\mathbf{G}$ for the $n$th training sample, $N$ is the number of training images, and $l^{SR}$ is a loss function that will be specified later.

For the discriminator network $\mathbf{D}$, $D(x)$ represents the probability of the data being the real high-resolution image rather than from $\mathbf{G}$. When training $\mathbf{D}$, we try to maximize its ability to differentiate ground-truth from the generated image, to force $\mathbf{G}$ to learn better details. When training $G$, we try to minimize $\log(1-D(G(\mathcal{T}_n, \theta_G), \theta_D))$, which is the log likelihood of $\mathbf{D}$ being able to tell that the image generated by $\mathbf{G}$ is not ground-truth. That is, we minimax the following function:
\begin{footnotesize}
\begin{eqnarray}
\mathop{\min}\limits_{\theta_G}\mathop{\max}\limits_{\theta_D}\mathbb{E}_{I^{HR}\sim p_{train}(I^{HR})}[\log(D(I^{HR}, \theta_D))]\\ \notag
\vspace{5em}+\mathbb{E}_{I^{HR}\sim p_{G}(\mathcal{T})}[\log(1-D(G(\mathcal{T}, \theta_G), \theta_D))].
\end{eqnarray}
\end{footnotesize}
In this way, we force the generator to optimize the generative loss, which is composed of perceptual loss, content loss and adversarial loss (more details of the loss function will be introduced in Section \ref{sec:training}).

\vspace{-0.3cm}
\subsubsection{Model architecture}
Our network is specialized for the analysis of time-series images through: (1) 3D filters in the neural network that take all the image frames into consideration, which extracts the time dependent information naturally, (2) two specifically designed modules in the generator residual network, i.e., Monte Carlo dropout \citep{mcdropout} and denoise shortcut, to cope with the stochastic switching of fluorophores and random noise, and (3) a novel incremental multi-scale architecture and parameter tuning scheme, which is designed to suppress the error accumulation in large upscaling factor neural networks.

\begin{figure}[!hpbt]
\centering
\vspace{-2em}
\includegraphics[width=0.38 \textwidth,type=png,ext=.png,read=.png]{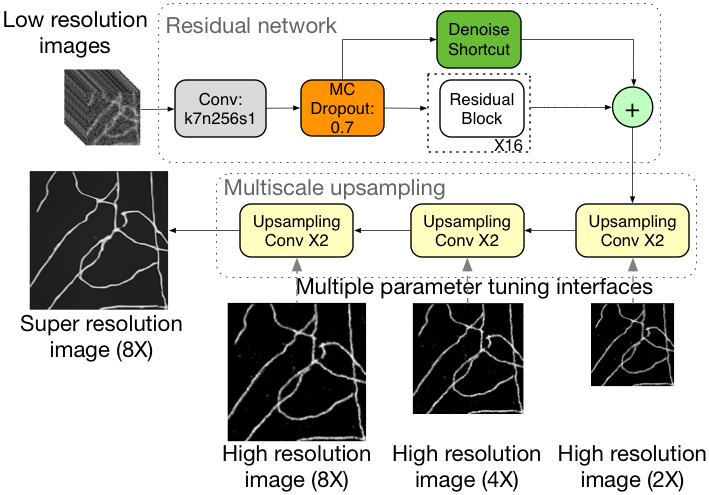}
\vspace{-1.5em}
\caption{Architecture of the generator network, which is composed of a residual network component and a multi-scale upsampling component. The low-resolution images are firstly fed to the residual network to extract information from the original 3D space, during which denoising is performed by the Monte Carlo dropout and denoise shortcut. The extracted feature maps are fed into the multi-scale upsampling component to increase the resolution gradually. We increase the resolution by a factor of 2 during each upsampling, resulting in three parameter tuning interfaces. Using the three interfaces, we can use all the $2\times$, $4\times$ and $8\times$ ground-truth images to train the generator network, reducing the artifacts in the final $8\times$ super-resolution images greatly.}
\label{fig:genor}
\vspace{-3.6em}
\end{figure}

Fig. \ref{fig:genor} illustrates the entire architecture of the generator model. The input is time-series low-resolution images. We first use a convolutional layer with the filter size as 7 by 7\footnote{Here, the filter size depends on the FWHM of a PSF. Generally, a fluorescence microscope produces low-resolution images with PSF spanning $3\sim7$ pixels. The specially designed filter size can balance between the computational time and physical meaning.}, which is larger than the commonly used filter, to capture meaningful features of the input fluorescence microscope images. The Monte Carlo dropout layer, which dropouts some pixels from the input feature maps during both training and testing, is applied to the output of the first layer to suppress noise. To further alleviate the noise issue, we use another technique, the denoise shortcut. It is similar to the identical shortcut in the residual network block. However, instead of being exactly the same as the input, we set each channel of the input feature map as the average of all the channels. The denoise shortcut is added to the output of the convolutional layer, which is after 16 residual blocks (Section S2), element-wise. After this feature map extraction process, we use a pixel shuffle layer combined with the convolutional layer to increase the dimensionality of the image gradually (upsampling Conv X2 in Fig. \ref{fig:genor}).

Here we adopt a novel multi-scale tuning procedure to stabilize the $8\times$ images. As shown in Fig. \ref{fig:genor}, our generator can output and thus calculate the training error of multi-scale super-resolution images, ranging from $2\times$ to $8\times$, which means that our model has multiple training interfaces for back propagation. Thus during training, we use the $2\times$, $4\times$, $8\times$ high-resolution ground-truth images to tune the model simultaneously to ensure that the dimensionality of the images increases smoothly and gradually without introducing too much fake detail.

For the discriminator network, we adopt the traditional convolutional neural network, which contains eight convolutional layers, one residual block and one sigmoid layer (Section S2). The convolutional layers increase the number of channels gradually to 2048 and then decrease it using 1 by 1 filters. Those convolutional layers are followed by a residual block, which further increases the model ability of extracting features. Section S2 provides detailed descriptions of the generator and discriminator networks.

\vspace{-0.3cm}
\subsubsection{Model training and testing} \label{sec:training}
GAN is known to be difficult to train \citep{salimans2016improved}. We use the following techniques to obtain stable models. For the generator model, we do not train GAN immediately after initialization. Instead, we pretrain the model. During the pretrain process, we minimize the mean squared error between the super-resolution image and the ground-truth, i.e., with the pixel-wise MSE loss as
\begin{footnotesize}
\begin{eqnarray}
  l^{SR}_{MSE_\mu} = \frac{1}{\mu^2WH}\sum_{x}^{\mu W}\sum_{y}^{\mu H}{(G(\mathcal{T}, \theta_{G_\mu})-I^{HR}_{x,y})^2},
\end{eqnarray}
\end{footnotesize}
\noindent where $W$ is the width of the low-resolution image, $H$ is the height of the low-resolution image, and $\mu=2,4,8$ is the upscaling factor. During pretraining, we optimize $l^{SR}_{MSE_8}$, $l^{SR}_{MSE_4}$, $l^{SR}_{MSE_2}$ simultaneously, instead of optimizing the sum of them.

Only after the model has been well-pretrained do we start training the GAN. During that process, we also use VGG19 \citep{simonyan2014very} to calculate the perceptual loss \citep{johnson2016perceptual} and use Adam optimizer \citep{kingma2014adam} with learning rate decay as the optimizer. When feeding an image to the VGG model, we resize the image to fulfill the dimensionality requirement:
\begin{footnotesize}
\begin{eqnarray}
  l^{SR}_{VGG_\mu} = \sum_{i=1}^{V}{(VGG(G(\mathcal{T}, \theta_{G_\mu}))_i-VGG(I^{HR})_i)^2},
\end{eqnarray}
\end{footnotesize}
where $V$ is the dimensionality of the VGG embedding output.

During final tuning, we simultaneously optimize the $2\times$, $4\times$, and $8\times$ upscaling by the generative loss:
\begin{footnotesize}
\begin{eqnarray}
  l^{SR}_{GAN_\mu} = 0.4*l^{SR}_{MSE_\mu} + 10^{-6}*l^{SR}_{VGG_\mu},
\end{eqnarray}
\end{footnotesize}
and
\begin{footnotesize}
\begin{eqnarray}
  l^{SR}_{GAN_8} = 0.5*l^{SR}_{MSE_8} + 10^{-3}*l^{SR}_{ADV_8} + 10^{-6}*l^{SR}_{VGG_8},
\end{eqnarray}
\end{footnotesize}
where $\mu=2,4$ and the $8\times$ upscaling has an additional term, the adversarial loss $l^{SR}_{ADV_8} = \sum_{n=1}^{N}{\log(1-D(G(\mathcal{T}_n, \theta_G), \theta_D))}$. For the discriminator network, we use the following loss function:
\begin{footnotesize}
\begin{eqnarray}
  l^{SR}_{DIS} = \sum_{n=1}^{N}{\log(D(G(\mathcal{T}_n, \theta_G), \theta_D))} + \sum_{n=1}^{N}{\log(1-D(I^{HR}_n, \theta_D))}.
\end{eqnarray}
\end{footnotesize}
\vspace{-0.5em}

During testing, for the same input time-series images, we run the model multiple times to get a series of super-resolution images. Because of the Monte Carlo dropout layer in the generator model, all of the super-resolution images are not identical. We then compute the average of these images as the final prediction, with another map showing the p-value of each pixel. We use Tensorflow combined with TensorLayer \citep{dong2017tensorlayer} to implement the deep learning module. Trained on a workstation with one Pascal Titan X, the model gets converged in around 8 hours.

\vspace{-0.3cm}
\subsection{The Bayesian inference module}
Our Bayesian inference module takes both the time-series low-resolution images and the primitive super-resolution image produced by the deep learning module as inputs, and generates a set of optimized fluorophore locations, which are further interpreted as a high-confident super-resolution image. Since the deep learning module has already depicted the ultrastructures in the image, we use these structures as the initialization of the fluorophore locations, re-sampling with a random punishment against artifacts. For each pixel, we re-sample the fluorophore intensity by $\sqrt{I_{x,y}}$ and the location by $(x,y)\pm rand(x,y)$, where $I_{x,y}$ is the pixel value in the image produced by deep learning, $rand(x,y)$ is limited in $\pm8$. In this way, the extremely high illumination can be suppressed and fake structures will be re-estimated.

\vspace{-0.3cm}
\subsubsection{Basic concepts}
As shown in Fig. \ref{fig:trans}, a fluorophore has three states: emitting (light), not emitting and bleached. In classic Bayesian-based time-series analysis, the switching procedure of fluorophores is modeled by Bayesian inference, i.e., given an observed region $R$, deciding whether there is a fluorophore ($F$) or not ($N$) by
\begin{footnotesize}
\begin{eqnarray}
  \frac{P(F|R)}{P(N|R)} = \frac{P(R|F)P(F)}{P(R|N)P(N)},	\label{eq:fn}
\end{eqnarray}
\end{footnotesize}
\noindent where $P(F)$ and $P(N)$ are constants which are based on experimental prior, $P(R|F)$ is the probability of the observed data region $R$ given the location of the fluorophore, $P(R|N)$ is the probability of the observed data region $R$ if there is no fluorophore, which can be calculated by integrating all the probability of observing pixels given the noise model.

\begin{figure}[!t]
\centering
\includegraphics[width=0.40 \textwidth,type=png,ext=.png,read=.png]{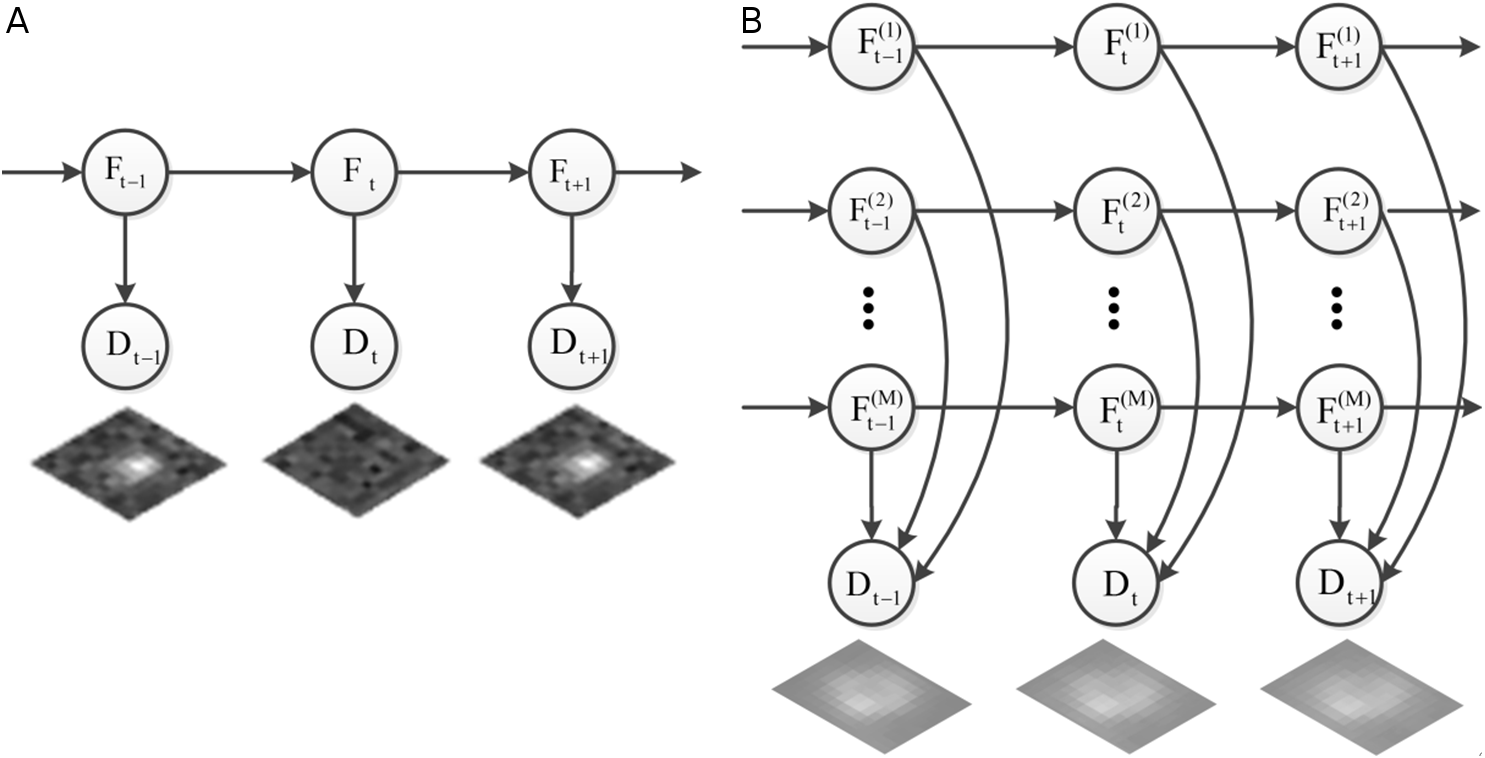}
\vspace{-1.5em}
\caption{The Bayesian inference model used for fluorophore switching. (A) For one fluorophore, its transition between different states can be modeled by a hidden Markov model (HMM). (B) For high-density fluorophores, the observed fluorescence and their underlying transitions can be modeled by a factorial hidden Markov model (FHMM).}
\label{fig:fhmm}
\vspace{-3.5em}
\end{figure}

For a single fluorophore, the switching procedure can be modeled by a hidden Markov model (HMM) \citep{rabiner1989tutorial}, as shown in Fig.\ref{fig:fhmm}(A). However, for high-density fluorophores, each fluorophore transfers the state independently with a stable probability \citep{cox2012bayesian} and all the fluorophores together can be modeled by a factorial hidden Markov model (FHMM) \citep{ghahramani1996factorial}, as shown in Fig.\ref{fig:fhmm}(B), which has been used and proved in \citep{xu2016live}.

\vspace{-0.3cm}
\subsubsection{Refining results with physical meaning}
Although HMM and FHMM are capable of modeling the fluorophore switching process, they are localization-guided, which often ignore the global information, and are computationally expensive to learn. Thus, we initialize the fluorophores' locations by using the image generated by deep learning and use Bayesian inference to further refine the results.

We apply the FHMM model to deal with high-density fluorescent microscopy. The parameters of FHMM are estimated by the expectation-maximization (EM) algorithm: 
\begin{footnotesize}
\begin{eqnarray}
  Q\left( {{\phi^{new}}|\phi} \right) = E\left\{ {\log P\left( {\left\{ {{F_t},{D_t}} \right\}|{\phi^{new}}} \right)|\phi,\left\{ {{D_t}} \right\}} \right\},	 \label{eq:fhmm-em}
\end{eqnarray}
\end{footnotesize}
\noindent where the observation sequence has $T$ frames, $\{D_t\}$, $t = 1,...,T$. The hidden states are $\{F_t\}$, where each fluorophore has three possible states in the model. $Q$ is a function of the fluorophore parameters $\phi^{new}$ given the current parameter estimation and the observation sequence $\{D_t\}$. The procedure iterates between a step that fixes the current parameters and computes posterior probabilities over the hidden states (the E-step), and a step that uses these probabilities to maximize the expected log likelihood of the observations as a function of the parameters (the M-step).

In the E-step, we fix the fluorophore parameters in the model and utilize the hybrid of Markov chain Monte Carlo and forward algorithm to sample the initial model. When a new fluorophore is determined, we take samples of this fluorophore using the forward filtering backward sampling algorithm \citep{godsill2004monte}. Thus, the sampled image sequence contains this fluorophore. In the M-step, we optimize the fluorophore parameters and find the maximum a posteriori (MAP) fluorophore positions using the conjugate gradient. Then, based on already known positions of fluorophores, the surrounding fluorophores with high probability are expended. The final super-resolution image is obtained by iterating these two steps until convergence.

The detailed method description and parameter setting are given in Section S3.

\vspace{-0.5cm}
\section{Experimental results}
\subsection{Training deep learning}
To train our deep learning module, the stochastic simulation module was used to simulate time-series low-resolution images from 12000 gray-scale high-resolution images. These images were downloaded from two databases: (i) 4000 natural images were downloaded from ILSVRC \citep{ILSVRC15} and Laplace filtered, and (ii) 8000 sketches were downloaded from the Sketchy Database \citep{sangkloy2016sketchy}. Note that our simulation is a generic method, which does not depend on the type of the input images. Thus any gray-scale image can be interpreted as the fluorophore distribution and used to generate the corresponding time-series low-resolution images.

To initialize all the weights of the deep learning models, we used the random normal initializer with the mean as $0$ and standard deviation as $0.02$. As for the Monte Carlo dropout layer, we set the keep ratio as $0.8$. In terms of the Adam optimizer \citep{kingma2014adam}, we followed the setting in \citep{RN140,Dai2017} and set the learning rate as $1*10^{-4}$ and the $beta\_1$, which is the exponential decay rate for the first moment estimates, as $0.9$. During training, we set the batch size as $8$, the initialization training epoch as $2$ and the GAN training epoch as $40$. When performing the real GAN training, we utilized the learning rate decay technique, reducing the learning rate by half every $10$ epochs.

\begin{figure*}[!hbpt]
\centering
\includegraphics[width=0.81\textwidth,type=png,ext=.png,read=.png]{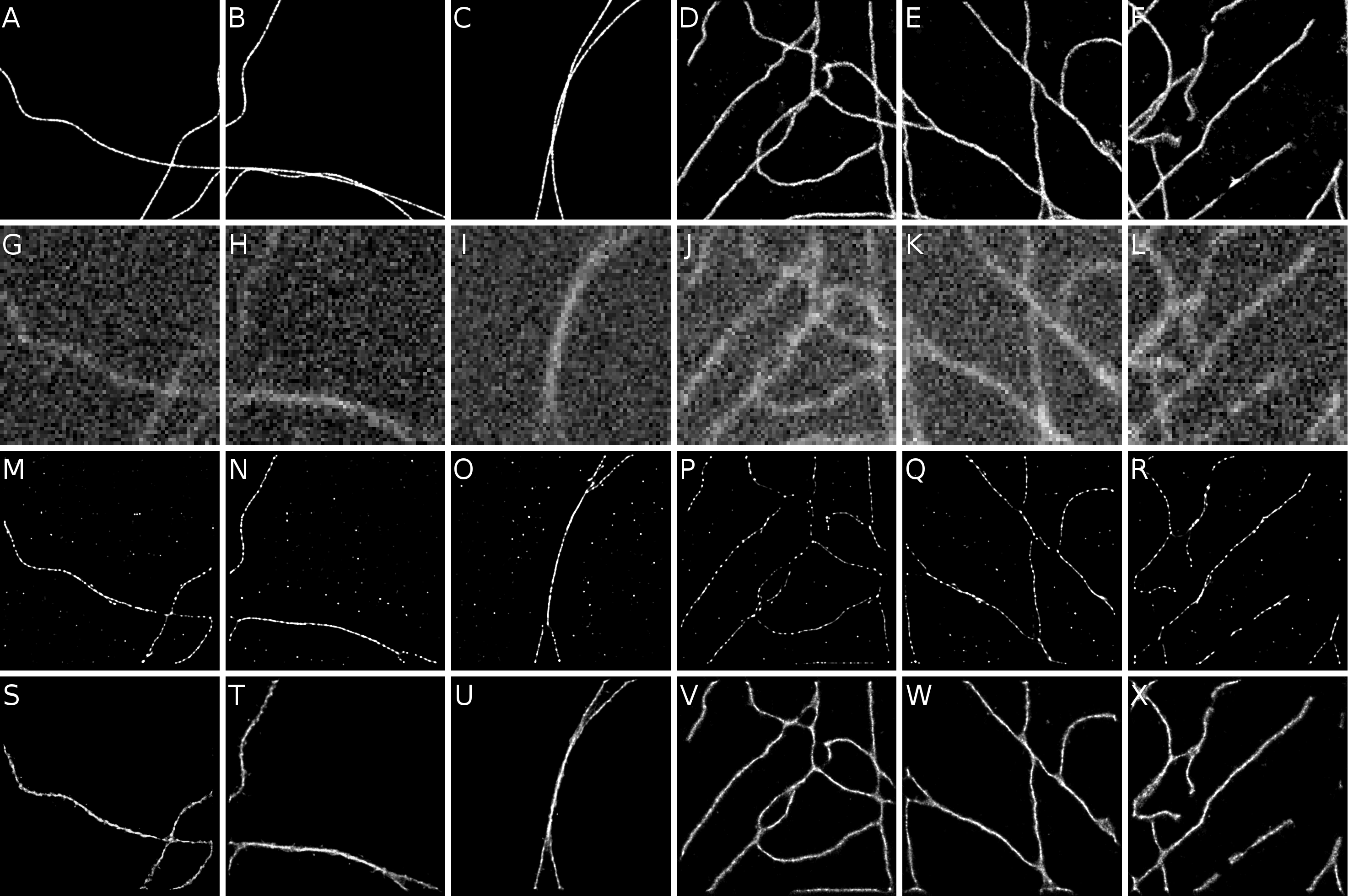}
\vspace{-1.5em}
\caption{Visualization of the ground-truth high-resolution images, representative low-resolution input images, the reconstruction results of the 3B analysis \citep{cox2012bayesian}, and the results of our method on three representative areas of each simulated dataset: MT (columns 1-3) and Tub (columns 4-6). The four rows show the ground-truth high-resolution images, the first frames of the simulated time-series low-resolution images, the reconstruction results of the 3B analysis, and the reconstruction results of DLBI, respectively.}
\vspace{-3.5em}
\label{fig:simu_result}
\end{figure*}

\vspace{-0.3cm}
\subsection{Evaluation datasets}
Two simulated datasets and three real-world datasets are used to evaluate the performance of the proposed method. Simulated datasets are used due to the availability of ground-truth.

The first two datasets are simulated datasets, for which the ground-truth (i.e., high-resolution images) is downloaded from the Single-Molecule Localization Microscopy (SMLM) challenge\footnote{http://bigwww.epfl.ch/smlm/datasets/} \citep{sage2015quantitative}. The two datasets correspond to two structures: MT0.N1.HD (abbr. MT) and Tubulin ConjAL647 (abbr. Tub). For each structure, single molecule positions were downloaded and then transformed to fluorophore densities according to Section 2.1. For simulation, the  photo-convertible fluorescent protein (PCFP) mEos3.2 \citep{zhang2012rational} and its associated PSF, FWHM and state transfer table were used. For the convenience of calculation, we cropped the large-field structure into four separate areas, each with $480\times480$ pixels ($1 px = 20 nm$). For each high-resolution image, 200 frames of low-resolution fluorescent images were generated, each with $60\times60$ pixels.

The third dataset is a real-world dataset, which was used in recent work \citep{xu2016live}. The actin was labeled with mEos3.2\footnote{For the convenience of cellular labeling and instrument setup, here all the experiments were carried out by mEos3.2.} in U2OS cells (abbr. Actin1) and \red{taken with an exposure time of 50 ms per image frame}. The actin network is highly dynamic and exhibits different subtype structures criss-crossing at various distances and angles, including stress fibers and bundles with different sizes and diameters. The dataset has 200 frames of high-density fluorescent images, each with $249\times395$ pixels ($1 px = 160 nm$) in the green channel. This is a good benchmark set that has been well tested which can compare our method with SIMBA \citep{xu2016live}, a recent Bayesian approach based on dual-channel imaging and photo-convertible fluorescent proteins.

Two other real-world datasets labeled with mEos3.2 were also used. One is an actin cytoskeleton network (abbr. Actin2), which is \red{labeled and taken under a similar exposure condition with} Actin1, but is completely new and has not been used by previous works. The other one is an Endoplasmic reticulum structure (abbr. ER), which has a more complex structure. It is a type of organelle that forms an interconnected network of flattened, membrane-enclosed sacs or tubes known as cisternae, which exhibits different circular-structures and connections at different scales. \red{For the ER dataset, the exposure time is 6.7 ms per frame.} The resolution of each image in Actin2 is $263\times337$ pixels ($1 px = 160 nm$) and that in ER is $256\times170$ pixels ($1 px = 100 nm$). Both datasets have 200 frames of high-density fluorescent images and the same photographing parameters as Actin1. These datasets were used to demonstrate the power of our method in diverse ultrastructures. The detailed procedure for collecting the real-world datasets is given in Section S4.

\red{Since the 3B analysis \citep{cox2012bayesian} is one of the most widely used high-density fluorescent super-resolution techniques, which can deal with high temporal and spatial resolutions \citep{lidke2012super,cox2012bayesian}, it was chosen to compare with our method.}

\vspace{-0.5cm}
\subsection{Performance on simulated datasets}

\subsubsection{Visual performance}

Fig. \ref{fig:simu_result} shows the visualization of the ground-truth high-resolution images, representative low-resolution input images, the reconstruction results of the 3B analysis, and the results of our method on the simulated datasets. Due to the space limitation, we illustrate three representative areas of each dataset and leave the fourth in Section S5.

As shown in Fig.\ref{fig:simu_result}, the ground-truth images have very clear structures while the low-resolution image frames are very blurry and noisy ($8\times$ downsampled). To reconstruct the ultrastructures, we ran the 3B analysis with 240 iterations and ran our Bayesian inference module after the deep learning module with 60 iterations. In each iteration, the Bayesian inference module of our method searches four neighbor points for each fluorophore, whereas the 3B analysis takes isolated estimation strategy. Thus the difference in iteration numbers is comparable. Due to the high computational expense of the 3B analysis, each $60\times60$ image was subdivided into nine overlapped subareas for multi-core process, whereas for our method, the entire image was processed by a single CPU core.

It is clear that the reconstructions of our method are very similar to the ground-truth in terms of smoothness, continuity, and thickness. On the other hand, the reconstructions of the 3B analysis consist of a number of interrupted short lines and points with thin structures. In general, two conclusions can be drawn from the visual inspection.

First, DLBI discovered much more natural structures than the 3B analysis. For example, in the bottom part of Fig. \ref{fig:simu_result}(B), there are two lines overlapping with each other and a bifurcation at the tail. Due to the very low resolution in the input time-series images (e.g., Fig. \ref{fig:simu_result}(H)), neither DLBI nor the 3B analysis was able to recover the overlapping structure. However, DLBI reconstructed the proper thickness of that structure (Fig. \ref{fig:simu_result}(T)), whereas the 3B analysis only recovered a very thin line structure (Fig. \ref{fig:simu_result}(N)). Moreover, the bifurcation structure was reconstructed naturally by DLBI. Similar conclusions can be drawn on the more complex structures in the Tub dataset (columns 4-6 in Fig. \ref{fig:simu_result}).

Second, DLBI discovered much more latent structures than the 3B analysis. The Tub dataset consists of a lot of lines (tubulins) with diverse curvature degrees (Fig. \ref{fig:simu_result}(D),(E),(F)). The reconstructions of the 3B analysis successfully revealed most of the tubulin structures but left the crossing parts interrupted (Fig. \ref{fig:simu_result}(P),(Q),(R)). As a comparison, the reconstruction results of DLBI recovered both the line-like tubulin structures and most of the crossing parts accurately (Fig. \ref{fig:simu_result}(V),(W),(X)).

\vspace{-0.3cm}
\subsubsection{Quantitative performance}
For single-molecule super-resolution fluorescence microscopy, the quantitative performance has been measured by assessing the localization accuracy of single-emitters in each frame \citep{ram2006beyond,small2009theoretical,huang2011simultaneous}. For high-density super-resolution fluorescence microscopy, the entire time-series are analyzed and the production is the probability map of the locations of fluorophores.

\begin{footnotesize}
\begin{table}[hbpt]
\vspace{-2.0em}
\centering
\captionsetup{font=scriptsize}
\caption{Performance comparison between the 3B analysis and DLBI on the four areas of the two simulated datasets in terms of peak signal-to-noise ratio (PSNR) and structural similarity (SSIM). The best performance is shown in bold.}
\vspace{-1em}
\scriptsize
\label{table:psnr}
\begin{tabular}{c|c|c c c c |c c c c }
\hline
\multicolumn{2}{c|}{\multirow{2}{*}{Datasets}} &\multicolumn{4}{c|}{MT0.N1.HD}&\multicolumn{4}{c}{Tubulin ConjAL647}       \\ \cline{3-10}  %
\multicolumn{2}{c|}{\multirow{2}{*}{}}  &   01   &   02   &   03   & \multicolumn{1}{c|}{04}  &   01   &   02   &   03   &   04    \\ \hline
PSNR&\small 3B & 17.99 & 17.62 & 17.84 & 17.89 & 13.42 & 15.49 & 15.00 & 13.21   \\
(dB)&\small DLBI & \textbf{18.59} & \textbf{19.16} & \textbf{18.51} & \textbf{20.42} & \textbf{18.72} & \textbf{19.17} & \textbf{18.72} & \textbf{16.63}  \\ \hline

SSIM&\small 3B &  0.89 &  0.89 &  0.90 & 0.90 &  0.74 &  0.81 &  0.75  & 0.69 \\
&\small DLBI &  \textbf{0.92} &  \textbf{0.92} &  \textbf{0.93} &   \textbf{0.94} &  \textbf{0.82} &  \textbf{0.85} &  \textbf{0.80} &  \textbf{0.76}    \\ \hline
\end{tabular}
\end{table}
\vspace{-2.em}
\end{footnotesize}

Since the ground-truth is known for the simulated datasets, here we use peak signal-to-noise ratio (PSNR) and structural similarity (SSIM) to measure the reconstruction performance, both of which are widely-used criteria for image reconstruction in computer vision. The performance of the 3B analysis and DLBI on the two simulated datasets are given in Table \ref{table:psnr}. Here, we denote the four areas (Section 3.2) of each dataset as ``01'', ``02'', ``03'' and ``04'', respectively. It can be seen that DLBI clearly outperforms the 3B analysis in terms of both PSNR and SSIM on all the areas of the two datasets.

\vspace{-0.4cm}
\subsection{Performance on real datasets}
\begin{figure}[!hbpt]
\centering
\vspace{-1.5em}
\includegraphics[width=0.30 \textwidth,type=png,ext=.png,read=.png]{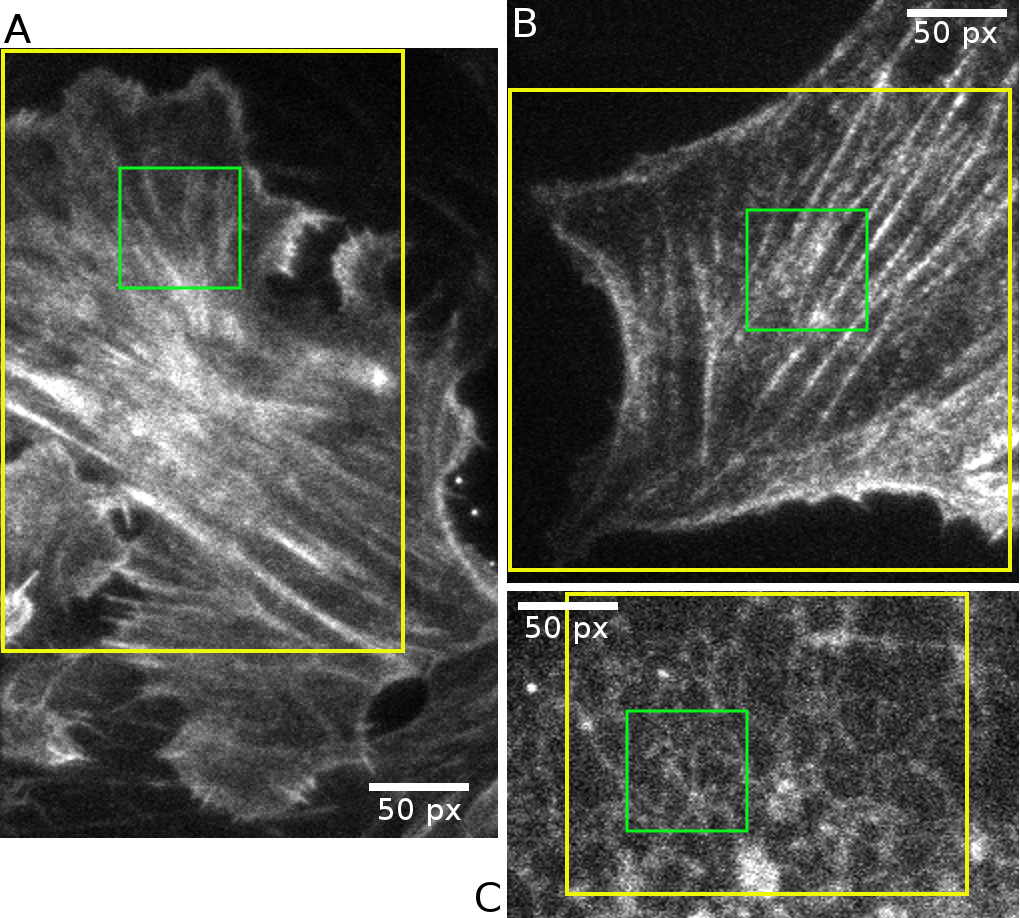}
\vspace{-1.3em}
\caption{The first frame of the time-series fluorescent images for each of the three real-world datasets: (A) Actin1 \citep{xu2016live}, (B) Actin2, and (C) ER. }
\label{fig:realdata}
\vspace{-5em}
\end{figure}

Fig. \ref{fig:realdata} shows the first frame of the time-series fluorescent images for each of the three real-world datasets. Here we evaluate the performance of our method for both local-patch reconstruction (areas selected by green rectangles) and large-field reconstruction (areas selected by yellow rectangles).

\subsubsection{Local-patch reconstruction}
\begin{figure}[!t]
\centering
\includegraphics[width=0.40\textwidth,type=png,ext=.png,read=.png]{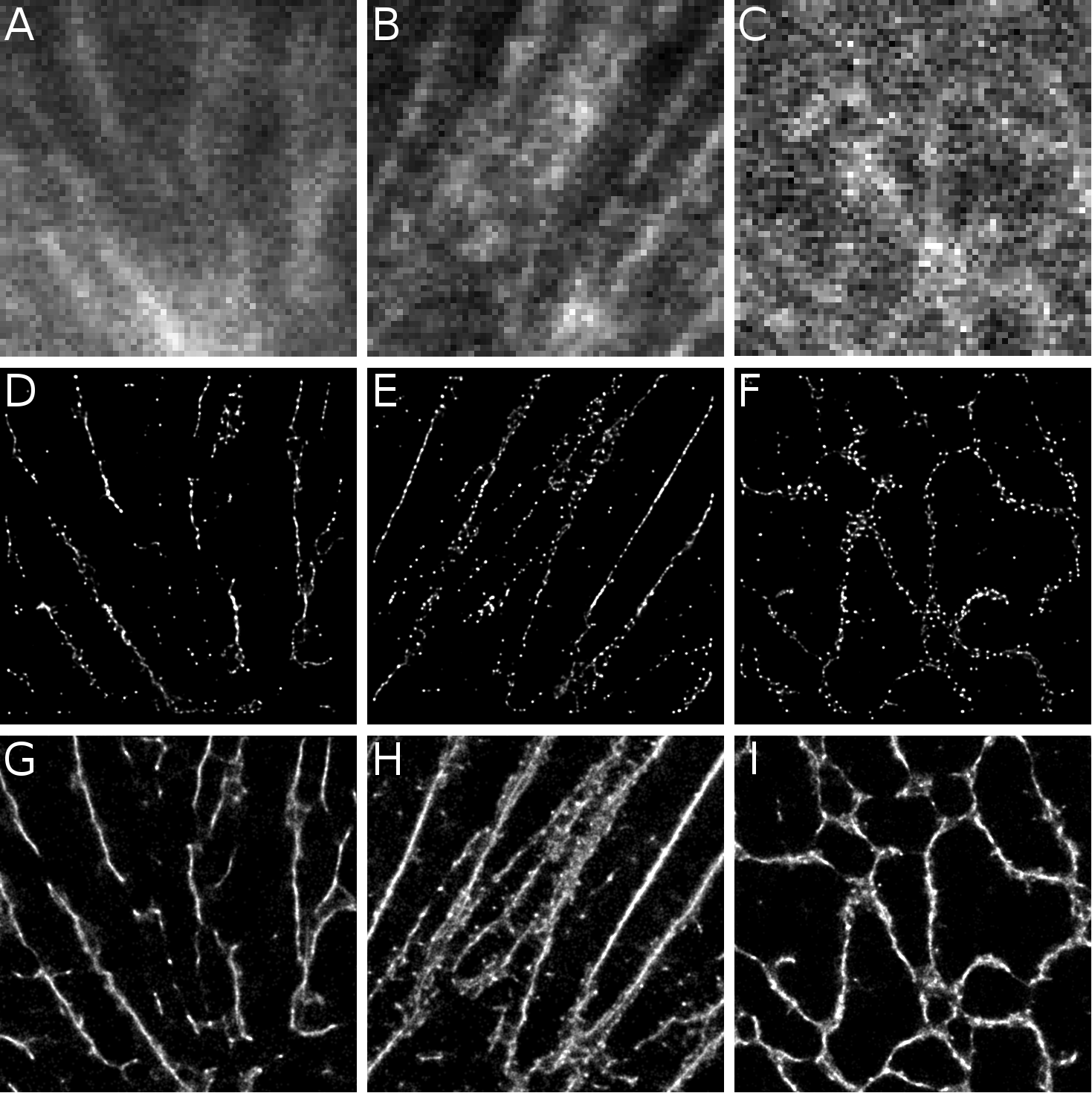}
\vspace{-1.5em}
\caption{Reconstructions of the local patches of the three real datasets. First column: the Actin1 dataset (the green box in Fig.\ref{fig:realdata}(A)). Second column: the Actin2 dataset (the green box in Fig.\ref{fig:realdata}(B)). Third column: the ER dataset (the green box in Fig.\ref{fig:realdata}(C)). The first row shows the first frames of the time-series low-resolution images. The second row shows the reconstructions of the 3B analysis. The third row shows the reconstructions of DLBI. The reconstructed images are $480\times480$ pixels and the local-patch images are $60\times60$ pixels.}
\vspace{-3.8em}
\label{fig:real_patch}
\end{figure}

Fig. \ref{fig:real_patch} shows the first frames of the low-resolution images of the three local-patches, and the reconstruction results of the 3B analysis and DLBI. The regions of interests of the selected patches are $(60,60,60,60)$, $(120,120,60,60)$ and $(60,60,60,60)$ for the three datasets, respectively\footnote{Region of interest is usually denoted as $(X,Y,W,H)$, where $(X,Y)$ are the coordinates of the top left point of the rectangle, $W$ is the width of the rectangle, and $H$ is the height of the rectangle.}. \red{The temporal resolutions for the two actin datasets and the ER dataset were 10s and 1.34s respectively, according to the exposure time of the image frames.}

It can be seen that the reconstruction results of the 3B analysis capture the main structures in the fluorescent images, but mainly consist of isolated high-illuminating spots, with details being interrupted (Fig. \ref{fig:real_patch}(D),(E),(F)). In contrast, the results of DLBI recover most of the latent ultrastructures, and the reconstructed structures have well-estimated fluorophore distribution and continuous depiction (Fig. \ref{fig:real_patch}(G),(H),(I)).

\begin{footnotesize}
\begin{table}[hbpt]
\vspace{-1.8em}
\centering
\captionsetup{font=scriptsize}
\caption{Performance comparison between the 3B analysis and DLBI on the real datasets in terms of RSP and RSE with SQUIRREL. The higher RSP and lower RSE values, the higher the image quality is. The best performance is shown in bold.}
\vspace{-1em}
\scriptsize
\label{table:squirrel}
\begin{tabular}{c|c c |c c |c c }
\hline
Dataset &\multicolumn{2}{c|}{Actin1-patch}&\multicolumn{2}{c|}{Actin2-patch}&\multicolumn{2}{c}{ER-patch}  \\ \hline
Criteria &   RSP   &   RSE &   RSP   &  RSE &   RSP   &   RSE        \\ \hline
 3B & 0.583 & 3915.096 & 0.770  & 2196.068 & 0.827 & 4077.037    \\
 DLBI & \textbf{0.721} & \textbf{3326.007} & \textbf{0.878} & \textbf{1648.919} & \textbf{0.916} & \textbf{2904.707}   \\ \hline
\end{tabular}
\end{table}
\vspace{-2em}
\end{footnotesize}
\red{We further assessed the reconstruction quality of the 3B analysis and DLBI by SQUIRREL (super-resolution quantitative image rating and reporting of error locations) \citep{culley2018quantitative}. SQUIRREL compares the diffraction-limited image (the reference image) and the reconstructed equivalents to generate a quantitative map, in which two scores are calculated: the resolution-scaled Pearson coefficient (RSP) and the resolution-scaled error (RSE). The higher RSP and lower RSE values, the higher the image quality is. Table \ref{table:squirrel} shows the RSP and RSE scores for the 3B analysis and DLBI. It is clear that DLBI significantly outperforms the 3B analysis. More detailed comparisons are given in Section S5.3. }

\vspace{-0.3cm}
\subsubsection{From deep learning to Bayesian inference}
Our method combines the strength of deep learning and statistical inference, where deep learning captures both local features in the images and the time-course correlation, and statistical inference removes artifacts from deep learning and enhances physical meaning to the final results. Conceptually, this is equivalent to using the power of deep learning to automatically and systematically explore and extract spatial and temporal features, and taking advantages of the explicit and rigorous mathematical foundation of probabilistic graphical models. Here we investigate the effectiveness of this combination.

\begin{figure}[!hbpt]
\centering
\vspace{-1.8em}
\includegraphics[width=0.38 \textwidth,type=png,ext=.png,read=.png]{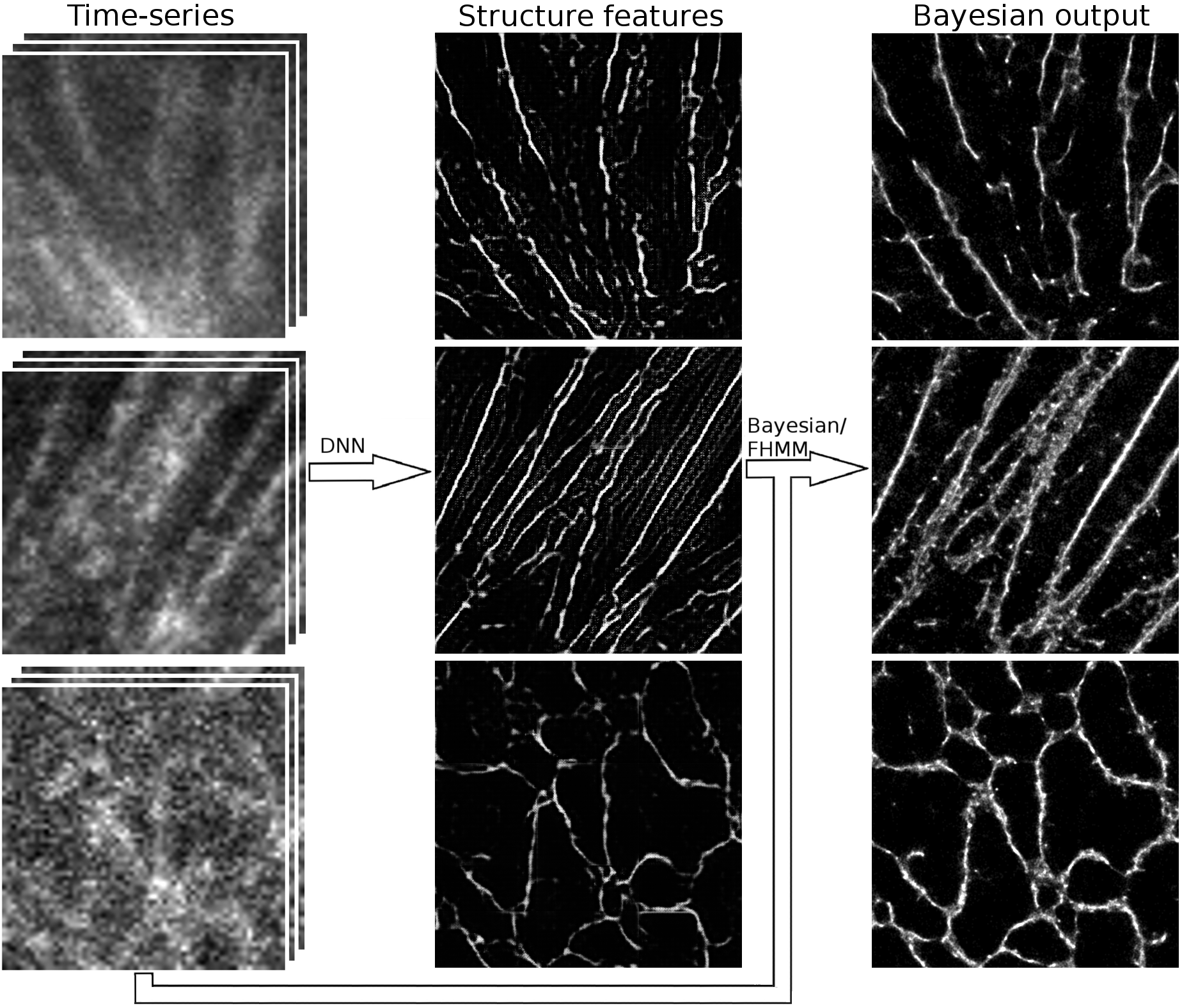}
\vspace{-1.5em}
\caption{Reconstructions of the three local patches of the three real datasets (the first column) by the deep learning module (the second column) and by deep learning guided Bayesian inference (the third column). }
\label{fig:dlbi}
\vspace{-3.5em}
\end{figure}

Fig. \ref{fig:dlbi} demonstrates the outputs of the deep learning module and the Bayesian inference module. It can be seen that the super-resolution images outputted from the deep learning module are very close to the final images from the Bayesian inference module, except for some artifacts and false structures. This is due to two reasons: (i) the abundance of training data provided by our simulation module, which are simulated under the real experimentally-calibrated parameters, enable deep learning to effectively learn spatial and temporal features; and (ii) the high diversity of biological structures is still a challenge, which causes the artifacts and false structures to be learned by deep learning.

After the deep learning module generates the super-resolution image, the Bayesian inference module uses both the original time-series low-resolution images and the deep learning image to statistically infer a ``false/true'' determination on each fluorophore location and produce the final image. In particular, the false structures are not directly rejected but used as seeds to search for true structures. Therefore, as shown in
Fig. \ref{fig:dlbi}, although the deep learning module outputted some unnatural structures for the Actin2 and ER datasets, these structures were further corrected by the Bayesian inference module.

\vspace{-0.3cm}
\subsubsection{Runtime analysis}
\begin{figure}[!hbpt]
\centering
\vspace{-1.8em}
\includegraphics[width=0.34\textwidth, height=0.22\textwidth,type=png,ext=.png,read=.png]{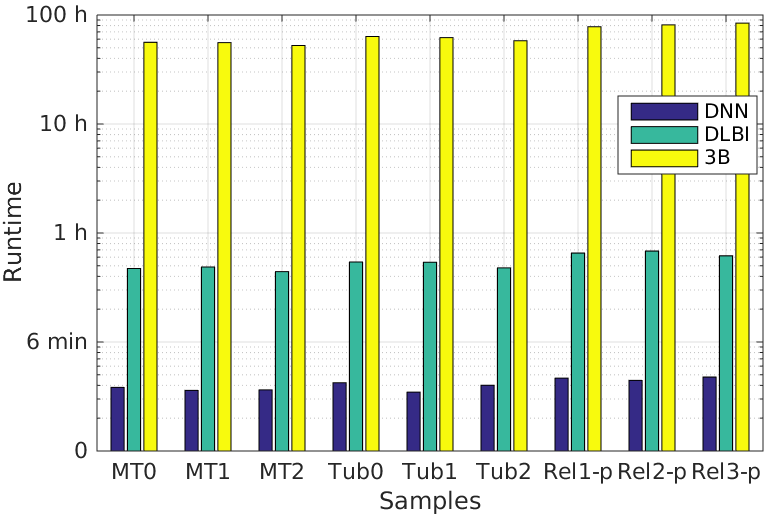}
\vspace{-1em}
\caption{Runtime comparison of the deep learning module (DNN), the entire DLBI pipeline (DLBI), and the 3B analysis (3B) on the nine reconstruction tasks (i.e., the six areas of the simulated datasets shown in Fig.\ref{fig:simu_result} and the three local patches of the real datasets shown in Fig.\ref{fig:real_patch}). The runtime was measured on a Fedora 25 system with 128 Gb memory and E5-2667v4 (3.2 GHz) CPU.}
\label{fig:runtime}
\vspace{-3.5em}
\end{figure}

After being trained, running the deep learning model is very computationally inexpensive. Furthermore, the results of deep learning provide a close-to-optimal initialization for Bayesian inference, which also significantly reduces trial-and-error and leads to faster convergence. Fig. \ref{fig:runtime} shows the runtime comparison of the deep learning module, the entire DLBI pipeline, and the 3B analysis on the nine reconstruction tasks (i.e., the six areas of the simulated datasets shown in Fig.\ref{fig:simu_result} and the three local patches of the real datasets shown in Fig.\ref{fig:real_patch}). It can be seen that the runtime for the deep learning module ranges between 1 to 3 minutes and that of DLBI ranges between 30 to 40 minutes. In contrast, the runtime for the 3B analysis is around 75 hours, which is more than 110 times higher than that for DLBI. Our results have demonstrated that the super-resolution images from the deep learning module alone is a good estimation to the ground-truth. Therefore, for users who value time and can compromise accuracy, the results from the deep learning module provide a good tradeoff, and thus a good estimation of the ground-truth.

\vspace{-0.3cm}
\subsubsection{Large-field reconstruction}
\begin{figure*}[!hbpt]
\centering
\includegraphics[width=0.75 \textwidth,type=png,ext=.png,read=.png]{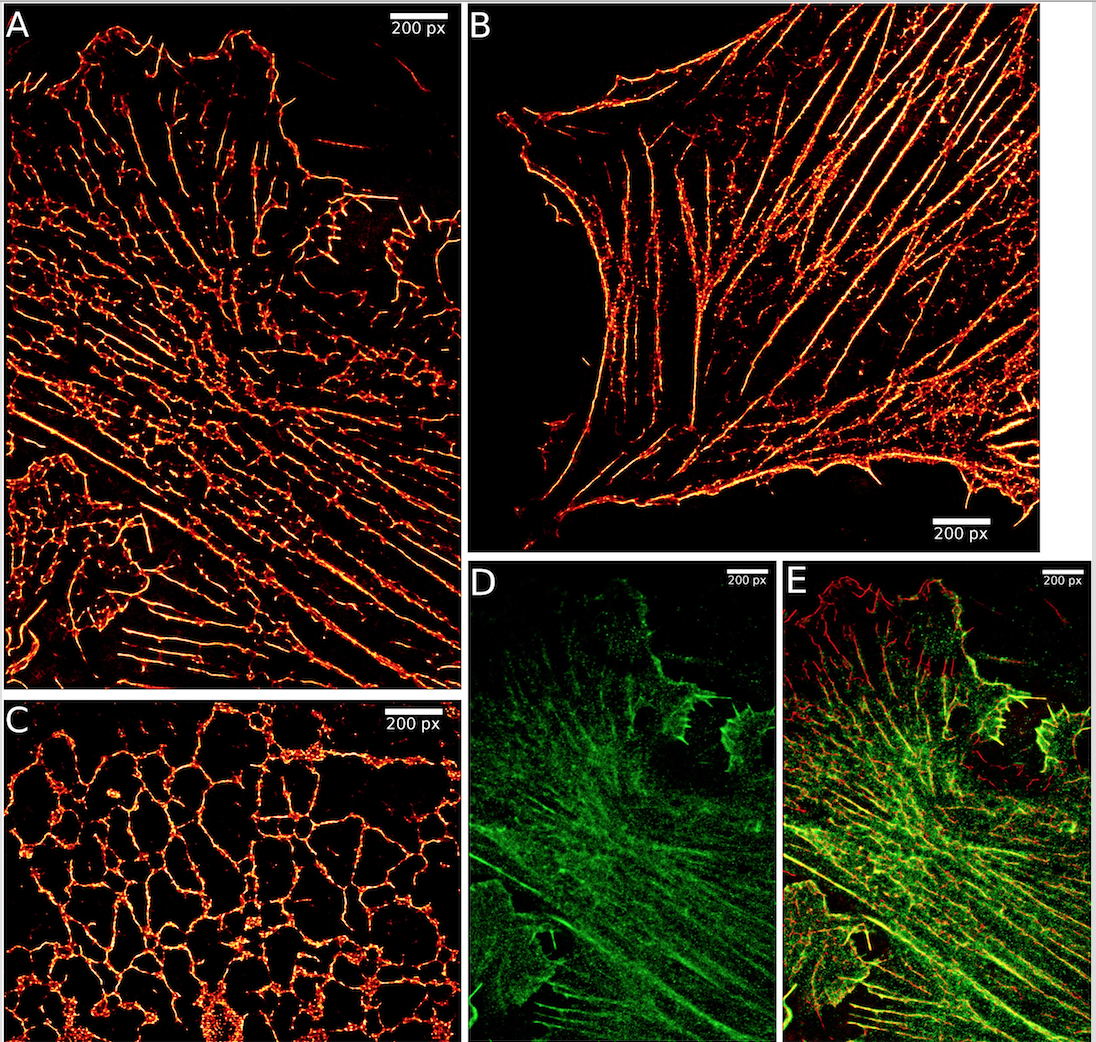}
\vspace{-1.3em}
\caption{Large-field reconstructions of the three real datasets: (A) Actin1, (B) Actin2, and (C) ER. (D) The single-molecule reconstruction of the Actin1 dataset by PALM based on 20,000 frames. (E) Overlap of the reconstruction images by DLBI (in red) and PALM (in green). }
\label{fig:wholescale}
\vspace{-3.5em}
\end{figure*}

To analyze a dataset with 200 frames, each with about $200\times300$ pixels, it takes our method about $7\sim10$ hours on a single CPU core. Therefore, our method is able to achieve large-field reconstruction on the yellow areas shown in Fig. \ref{fig:realdata}. Fig. \ref{fig:wholescale} shows the large-field reconstruction images of the three real datasets. For the Actin1 dataset, the selected area is $200\times300$ pixels and the reconstructed super-resolution image is $1600\times2400$ pixels. For the Actin2 dataset, the selected area is $250\times240$ pixels and the reconstructed image is $2000\times1920$ pixels. And for the ER dataset, the selected area is $200\times150$ pixels and the reconstructed image is $1600\times1200$ pixels.

As shown in Fig. \ref{fig:wholescale}(A) and (B), the actin networks in the two datasets have been successfully recovered by DLBI. The thinning and thickening trends of the cytoskeleton have been clearly depicted, as well as the small latent structures, including actin filaments, actin bundles and ruffles. For the endoplasmic reticulum structure (Fig.\ref{fig:wholescale}(C)), the circular-structures and connections of the cytoskeleton have also been accurately reconstructed.

For the Actin1 dataset, the single-molecule reconstruction of the red channel is available (Fig. \ref{fig:wholescale}(D)). This reconstruction was produced by PALM \citep{hess2006ultra} using 20,000 frames, whereas the reconstruction image of DLBI (Fig. \ref{fig:wholescale}(A)) used only 200 frames. We further overlayed the image produced by DLBI with that of PALM to check how well they overlap (Fig. \ref{fig:wholescale}(E)). It is clear that the main structures of the two images almost perfectly agree with each other. In addition, our method was able to recover the latent structure on the top-left part which was not photographed by PALM \red{due to out of range of views in dual-channel photographing}. If we carefully check the original low-resolution fluorescent images, we could find that this predicted structure indeed exists, which is consistent with our reconstruction.

\vspace{-0.7cm}
\section{Discussion and conclusion}
In this paper, we proposed a deep learning guided Bayesian inference method for structure reconstruction of super-resolution fluorescence microscopy. Our method combines the strength of deep learning and statistical inference. We further overcame the high data requirement bottleneck of deep learning by a novel stochastic simulation module based on the experimentally-calibrated parameters and problem-specific physical models.
It should be noted that although our simulation provides close-to-realistic data to train the deep learning module, it still contains bias and unrealistic parts, which will be learned by the deep learning module. Bayesian inference, on the other hand, can correct and refine the ultrastructure learned by deep learning, and thus enhance the physical meaning of the final super-resolution image.

\red{We have comprehensively evaluated the quality of the reconstructed super-resolution images. The future work includes evaluating how well the framework can rediscover the parameters used in the simulation module.}

\vspace{-0.7cm}
\section*{Acknowledgements}
\red{This work was supported by the Kind Abdullah Unviersity of Science and Technology (KAUST) Office of Sponsored Research (OSR) under Awards No. FCC/1/1976-04, URF/1/2602-01, URF/1/3007-01, URF/1/3412-01 and URF/1/3450-01, the National Key Reaseach and Development Program of China (2017YFA0504702), the National natural Science Foundation of China (Grant No. U1611263, U1611261, 61232001, 61472397, 61502455, 61672493).}

\vspace{-0.7cm}
\bibliographystyle{natbib}
\bibliography{reference}

\appendix

\end{document}